This is a preprint copy that has been accepted for publication in *International Journal of Electrical Power and Energy Systems*.

Please cite this article as:

Tao Xiong, Yukun Bao, Zhongyi Hu. "Interval Forecasting of Electricity Demand: A Novel Bivariate EMD-based Support Vector Regression Modeling Framework", *International Journal of Electrical Power and Energy Systems*. 2014, 10.1016/j.ijepes.2014.06.010 . Accepted

# Research Highlights

➢ Proposing a novel interval-valued electricity demand forecasting approach.

➢ BEMD and SVR are integrated for interval forecasting of electricity demand.

➢ The EMD-based modeling framework are extended to deal with interval forecasting

➢ BEMD is used to decompose both the lower and upper bounds electricity demand series.

➢ The proposed modeling framework is justified with real world data sets.

# Interval Forecasting of Electricity Demand: A Novel Bivariate

# EMD-based Support Vector Regression Modeling Framework

Tao Xiong[1], Yukun Bao[*][2], Zhongyi Hu[2]

[1]   College of Economics and Management, Huazhong Agricultural University, Wuhan, P.R.China, 630027

[2] School of Management, Huazhong University of Science and Technology, Wuhan, P.R.China, 430074

## Abstract

Highly accurate interval forecasting of electricity demand is fundamental to the success of reducing the risk when making power system planning and operational decisions by providing a range rather than point estimation. In this study, a novel modeling framework integrating bivariate empirical mode decomposition (BEMD) and support vector regression (SVR), extended from the well-established empirical mode decomposition (EMD) based time series modeling framework in the energy demand forecasting literature, is proposed for interval forecasting of electricity demand. The novelty of this study arises from the employment of BEMD, a new extension of classical empirical model decomposition (EMD) destined to handle bivariate time series treated as complex-valued time series, as decomposition method instead of classical EMD only capable of decomposing one-dimensional single-valued time series. This proposed modeling framework is endowed with BEMD to decompose simultaneously both the lower and upper bounds time series, constructed in forms of complex-valued time series, of electricity demand on a monthly per hour basis, resulting in capturing the potential interrelationship between lower and upper bounds. The proposed modeling framework is justified with monthly interval-valued electricity demand data per hour in Pennsylvania–New Jersey–Maryland Interconnection, indicating it as a promising method for interval-valued electricity demand forecasting.

[*] Corresponding author: Tel: +86-27-87558579; fax: +86-27-87556437.
Email: yukunbao@hust.edu.cn or y.bao@ieee.org

## 1. Introduction

To supply high quality electric energy to the customer in a secure and economic manner, an electricity operator faces many technical and economical problems in operation, planning, control and reliability management of an electric power system. In achieving this goal, accurate forecast of electricity demand is the first prerequisite.

According to extent literature investigation, it is not hard to find that a wide variety of methodologies and techniques have been used for electricity demand forecasting with many degrees of success, such as exponential smoothing models [1], regression models [2], fuzzy logic approach [3], fuzzy inference system [4], grey-based approaches [5], wavelet transforms and adaptive models [6], kernel-based method [7], artificial neural networks [8, 9], support vector machines [10], semi-parametric method [11] and hybrid model [12, 13]. The reader is referred to [14] for a recent survey of the presented methodologies and techniques employed for electricity demand forecasting. However, an important point to note from past studies mentioned above is their preoccupation with point forecasting rather than interval one.

An interval forecasting of electricity demand has the advantage of taking into account the variability and/or uncertainty so as to reduce the amount of random variation relative to that found in classic single-valued load time series. As García-Ascanio and Maté [15] pointed out, the interval-valued time series forecasting (ITS) methods as a potential tool that will lead to a reduction in risk when making power system planning and operational decisions. To date, several suitable tools for managing ITS have been developed (see [16] for a recently review), such as interval

Holt's exponential smoothing methods [17], interval multi-layer perceptrons (iMLP) [18], vector autoregressive (VAR) model [15], vector error correction (VEC) model [19], etc. Most of the conventional methodologies available for ITS in the literature propose the use of computational methods or modeling schema, accounting for the capability of dealing with interval-valued data. For instance, neural networks are an area that has recently witnessed substantial improvements in interval analysis. Notable earlier work on interval analysis using neural networks includes that of Simoff [20], while Beheshti, Berrached [21] developed a multi-layer perceptron (MLP) in which inputs, weights, biases, and outputs are intervals, and proposed an algorithm so as to obtain the optimal weights and biases. Recently, Roque, Maté [18] proposed and analyzed a new model of MLP (named iMLP in [18]) based on interval arithmetic that facilitates handling input and output interval-valued data, but where weights and biases are single-valued and not interval-valued.

Our study focuses on extending the EMD-based time series modeling framework to adapt to the scenario of interval forecasting of electricity demand. Following the philosophy of 'divide and conquer', EMD-based time series modeling framework has been recently well-established and well-justified in energy market [22-25], tourism management [26], hydrology [27], and transportation research [28]. Generally speaking, there are three main steps involved in the classical EMD-based time series modeling framework, i.e., decomposition, single forecast, and ensemble forecast. First of all, EMD is used to divide the original single-valued time series into a finite number of intrinsic mode function (IMFs) components and a residue component.

Secondly, some powerful modeling techniques such as artificial neural network (ANN) and support vector regression (SVR), are applied to model and predict all components (IMFs and residue) respectively. Finally, these prediction results of all extracted IMFs components and the residue in the previous step are combined to generate an aggregated output using one modeling technique, which can be seen as the final prediction results for the original time series. However, the classical EMD applied in the aforementioned studies [22-28] is only applicable to one-dimensional (univarite) single-value time series decomposition. One straightforward solution to extend the classical EMD-based modeling framework for interval forecasting of electricity demand is to decompose and forecast the lower and upper bound series of interval-valued electricity demand respectively as several studies [29, 30] do, without considering the possible interrelations that are presented amongst themselves, which has been criticized in [16]. To enhance the capability of classical EMD, fortunately, BEMD [31] has been recently proposed to extend this decomposition method to treat complex-valued signals. Although BEMD was not specially developed for interval-valued time series analysis purposes, in view of the BEMD's advantages in decomposing complex-valued signals, this study proposes to construct a interval-valued electricity demand time series consisting of both lower and upper bounds of monthly electricity demand per hour in forms of complex-valued time series. This is to say, the lower and upper bounds series are taken as the real and imaginary parts of the complex-valued signal respectively. After this generic construction, the interval-valued time series of electricity demand can fully be decomposed through BEMD and then be fed into the proposed BEMD-based time series modeling framework for forecasting task after the selection of a specific modeling technique such as SVR in this study.

Our contributions could be outlined as follows. First, as was mentioned above, although extensive mounts of approaches [2-14] have been developed for electricity demand modeling and forecasting, most of them rely only on single-valued electricity demand series. The interval forecasting of electricity demand has not been widely explored (In fact, we have only come across interval-valued electricity demand forecasting in one published work [15]). Second, by introducing BEMD, the well-established EMD-based time series modeling framework can be extended to deal with interval-valued time series forecasting, which can fully take the advantages of its simplification of modeling process in nature of 'divide and conquer' as well as avoiding to increase computational cost while employing complex-valued modeling techniques to deal with interval-valued time series. Third, since the work of Rilling, Flandrin [31], BEMD has attracted particular attention in engineering technology filed [32-35]. However, there have been very few, if any, studies for interval-valued time series forecasting using the BEMD-based modeling framework. So, we hope this study would fill this gap. The fourth contribution is straightforward to provide the empirical evidence on the interval-valued electricity demand forecasting with real-world data from Pennsylvania–New Jersey–Maryland (PJM) market. Given the hourly single-valued electricity demand series from PJM market, we calculate the maximum and minimum value of the demand per hour and month from 2000 to 2011.

This produces an interval-valued time series where each observation is formed by an interval that collects, as its lower bound, the minimum value of the electricity demand and, as its upper bound, the maximum value of the electricity demand for a specific hour, month and year.

This paper is structured as follows. In Section 2, we provide brief introduction of BEMD and illustrate the data representation of interval-valued electricity demand time series analysis. Afterwards, the proposed BEMD-based SVR modeling framework are discussed in detail in Section 3. Section 4 details the research design on data process and preliminary analysis, accuracy measure, methodologies implementation, and experimental procedure. Following that, in Section 5, the experimental results are discussed. Section 6 finally concludes this work.

## 2. BEMD with interval-valued electricity demand time series

In this section, the overall formulation process of the BEMD for interval-valued electricity demand time series forecasting is presented. First, the data representation of interval-valued electricity demand time series is illustrated. Then the BEMD for the obtained ITS is formulated in details.

### 2.1 Constructing interval-valued electricity demand time series

Classical statistics and data analysis deal with individuals who can be described by a classic variable that takes as its value either a real value (for a quantitative variable) or a category (for a nominal variable). However, observations and estimations in the real world are usually incomplete to represent classic data exactly. In the electricity market, for instance, electricity demand has its daily (or weekly, or monthly) bounds and varies in each period-day, week, or month. Representing the variations with snap shot points, say the highest demand, only reflects a particular number at a particular time; it does not properly reflect its variability during the period. This problem can be reduced if the higher and lower demand per period is considered, giving rise to an interval-valued time series.

Interval-valued data is a particular case of symbolic data in the field of symbolic data analysis (SDA) [36]. SDA states that symbolic variables (lists, intervals, frequency distributions, etc) are better suited than single-valued variables for faithfully describing complex real-life situations [30]. It should be noted that interval-valued data in the field of SDA do not come from noise assumptions, but rather from the expression of variation or aggregation of huge databases into a reduced number of groups [17].

In the context of SDA, an interval-valued variable, $[Y]$, is a variable defined for all the elements $i$ of a set $E$, where $[Y_i] = \left\{ \left[ Y_i^L, Y_i^U \right] : Y_i^L, Y_i^U \in \mathbb{R}, Y_i^L \leq Y_i^U \right\}, \forall i \in E$. The particular value of $[Y]$ for the $i$th element can be either denoted by the interval lower and upper bounds $[Y_i] = \left[ Y_i^L, Y_i^U \right]$ or the center (mid-point) and radius (half-range) $[Y_i] = \left[ Y_i^C, Y_i^R \right]$, where $Y_i^C = \left( Y_i^L + Y_i^U \right) \big/ 2$ and $Y_i^R = \left( Y_i^U - Y_i^L \right) \big/ 2$. In Table 1, the interval-valued in every month of the hourly spot electricity demand in Pennsylvania–New Jersey–Maryland (PJM) Interconnection in MWh per day and per hour in 2011 is showed.

**<Insert Table 1 here>**

An interval-valued time series (ITS) is a chronological sequence of interval-valued variable, the value of the variable in each instant of time $t$ $(t = 1, \ldots, n)$ is expressed as a two-dimensional vector $\left[ Y_t^L, Y_t^U \right]$ with the elements in $\mathbb{R}$ representing the lower bound $Y_t^L$ and upper bound $Y_t^U$, with $Y_t^L \leq Y_t^U$. Thus, an ITS is $\left[ Y_t \right] = \left[ Y_t^L, Y_t^U \right]$ for $t = 1, \ldots, n$, where $n$ denotes the number of intervals of the time series (sample size).

Fig. 1 illustrates the electricity market in which a monthly interval-valued electricity demand series in 2011 for the hour 10, H10, arises. Fig.1(a) illustrates a daily electricity demand series for H10, 2011. Fig. 1(b) depicts the corresponding monthly electricity demand intervals, which are obtained at each month from the minimal and maximal values of the electricity demand time series at the Fig. 1(a).

**<Insert Fig.1 here>**

## 2.2 Decomposing with BMED

Empirical mode decomposition (EMD), first proposed by Huang, Shen [37], is an empirical, intuitive, direct and self-adaptive time series analysis tool, with which any signals can be decomposed into a finite number of independent and nearly periodic intrinsic mode function (IMFs) components and a residue based purely on the local characteristic time scale. Ensemble empirical mode decomposition (EEMD) proposed by Wu and Huang [38] is an improved version of EMD. It is proposed to overcome intrinsic drawbacks of mode mixing in EMD.

Since the seminal works of Huang, Shen [37] and Wu and Huang [38] were published, EMD and EEMD have been widely used for time series analysis, which indicating the superiority EMD/EEMD-based modeling framework for time series forecasting by decomposing single-valued time series using EMD/EEMD, particularly in the energy market analysis [22-25]. However, traditional EMD/EEMD is only suitable for one-dimensional single-valued signals. Currently, the most common extensions of EMD to the field of complex numbers have rely on bivariate EMD [31] which operate directly in complex space $\mathbb{C}$ making it suitable in practical applications [39]. Hence, BEMD is adopted in this study for simultaneously processing interval-valued electricity demand data based on the intuition that the lower and upper bounds of monthly electricity demand per hour do not drift apart over time.

It is worth to note that BEMD was not specially developed for interval-valued time series analysis purposes. Hence, we first construct the complex-valued signal suitably processed by BEMD using the considered interval-valued electricity demand series. Given an interval-valued electricity demand series $\left[ Y_t \right] = \left[ Y_t^L, Y_t^U \right]$ for $t = 1, \ldots, n$ as shown in Fig. 2, where $n$ denotes the number of intervals of the time series and $Y_t \in \mathbb{R}^2$ represents the $t$th interval. The data are expressed in log scale. We need a mapping $\mathbb{R}^2 \rightarrow \mathbb{C}$ to process the information with BEMD. One such mapping for each interval of the interval-valued electricity demand series can be done by the following transformation:

$$C_t = Y_t^L + i * Y_t^U, t = 1, \ldots, n \tag{1}$$

or

$$C_t = Y_t^U + i * Y_t^L, t = 1, \ldots, n \tag{2}$$

where $C_t\left(t=1,\ldots,n\right)$ is the resulting complex-valued signal, $Y_t^L$ and $Y_t^U$ $\left(t=1,\ldots,n\right)$ are the lower and upper bounds of considered interval-valued electricity demand series, respectively, and $i=\sqrt{-1}$. In other words, a complex-valued signal can be constructed by considering the lower and upper bounds of interval-valued signal as the real (imaginary) and imaginary (real) parts of the complex-valued signal, respectively. This kind of complex-valued signals construction has been widely used in the field of signal processing, e.g., real-valued fast Fourier transform algorithms [40]. Afterwards, the BEMD is applied on both the lower and upper bounds simultaneously because interval-valued electricity demand have a mutual dependence, e.g., cointegration (the empirical evidences are given in Section 4.1), between the lower and upper bounds.

**\<Insert Fig.2 here\>**

It should be noted that which of two transformations (Eq. (1) and Eq. (2)) is better is still an unresolved question. In this study, we attempt to compare these two transformations with empirical evidence in the context of interval forecasting of electricity demand. For the purpose of illustration, the transformation Eq. (1) is raised as an example in the following sections.

Fig. 3 presents the BEMD results for the aforementioned interval-valued electricity demand. In Fig. 3, the lower and upper bounds of considered interval-valued electricity demand is decomposed into five IMFs and a residue, respectively.

**\<Insert Fig.3 here\>**

Detailed discussions on the BEMD can be found in [31], but a brief introduction about formulation is provided here. BEMD is similar to EMD in calculation except in modifications of extrema detection and envelop definition. Its algorithm can be described as follows [31-33].

Given a complex-valued signal $C_t = Y_t^L + i*Y_t^U$ for $t = 1,\ldots,n$, where $Y_t^L$ and $Y_t^U$ represent the lower and upper bounds, respectively, and $i = \sqrt{-1}$.

*Step* 1: Let $\tilde{C}_t = C_t$.

*Step* 2: Project the complex-valued signal $\tilde{C}_t$ on directions $\varphi_m$

$$p_t^{\varphi_m} = R\left\{e^{-j\varphi_m}\tilde{C}_t\right\}, \quad m \in [1, M] \tag{3}$$

where $R\{\cdot\}$ denotes the real part of a complex-valued signal, $\varphi_m = 2m\pi/N$ denotes projection directions, $M$ denotes the number of projection directions, and $e^{-j\varphi_m}$ denotes a unit complex number.

*Step* 3: Find the locations $\left\{\left(t_i^m, p_{t_i^m}^{\varphi_m}\right)\right\}$ corresponding to the local maxima of $p_t^{\varphi_m}$.

*Step* 4: Interpolate (using spline interpolation) the set $\left\{\left(t_i^m, p_{t_i^m}^{\varphi_m}\right)\right\}$ to obtain the envelop curve $e_t^{\varphi_m}$.

*Step* 5: Compute the mean of all envelop curves $o_t = \frac{1}{M}\sum_{m=1}^{M}e_t^{\varphi_m}$ and subtract from the input signal, that is, $h_t = \tilde{C}_t - o_t$. Let $\tilde{C}_t = h_t$, and go to *Step* 2. Repeat until $h_t$ satisfies the physical sense of an IMF (see the definition of an IMF from [37])

*Step* 6: Record the obtained IMF and remove it from $C_t$, i.e.

$$g_t^1 = h_t \tag{4}$$

where $g_t^1$ denotes the first IMF.

$$r_t = C_t - h_t \tag{5}$$

where $r_t$ denotes the residual.

*Step* 7: Repeat above procedure to the residual $r_t$ to obtain all the IMFs.

*Step* 8: Once all IMFs contained in $C_t$ have been obtained, the original complex-valued $C_t$ may be expressed as

$$C_t = \sum_{i=1}^{I} g_t^i + r_t \qquad (6)$$

where $I$ indicates the total number of IMFs, $g_t^i, i = 1, 2, \ldots, I$ denote the IMFs, and $r_t$ denotes the residual.

## 3. The proposed BEMD-SVR modeling framework

In this section, the proposed BEMD-based SVR modeling framework is formulated and corresponding steps involved in this implementation are presented in details.

Given there is an interval-valued electricity demand series $\left[Y_t\right] = \left[Y_t^L, Y_t^U\right]$ for $t = 1, \ldots, n$, where $Y_t^L$ and $Y_t^U$ represent the lower and upper bound of electricity demand at time $t$, respectively. A four-steps modeling framework integrating BEMD and SVR can be formulated for interval-valued electricity demand forecasting (Fig. 4).

As shown in Fig. 4, the proposed BEMD-based SVR modeling framework is generally composed of the following four main steps:

*Step* 1: Complex-valued signal construction. The original interval-valued electricity demand $\left[Y_t\right] = \left[Y_t^L, Y_t^U\right]$ for $t = 1, \ldots, n$ is transformed into a complex-valued signal $C_t = Y_t^L + i * Y_t^U$ for $t = 1, \ldots, n$ by considering the lower and upper bounds as the real and imaginary parts of the complex-valued signal, respectively, where $i = \sqrt{-1}$ .

*Step* 2: Decomposition. The resulting complex-valued signal $C_t \left( t = 1, \ldots, n \right)$ is decomposed into $I$ complex-valued intrinsic mode functions (IMFs) components $g_t^i, i = 1, 2, \ldots, I$ , and one complex-valued residual component $r_t$ using BEMD algorithm. Accordingly, the real and imaginary parts of the obtained complex-valued components, i.e., $g_t^i, i = 1, 2, \ldots, I$ , and $r_t$ , are the extracted components (IMFs and residual) of lower and upper bounds series, respectively.

*Step* 3: Single forecast: SVR is used as single forecasting tool to independently model the extracted components (IMFs and residual) of the lower and upper bounds series, respectively. Accordingly, the corresponding prediction results for all components of lower and upper bounds series can be obtained, respectively.

*Step* 4: Ensemble forecast: Prediction results of all components (IMFs and residual) of lower and upper bounds series produced by SVR in the previous step are combined to generate an aggregated output, respectively, which can be seen as the final prediction of lower and upper bounds for the original interval-valued electricity demand series $\left[ Y_t \right] = \left[ Y_t^L, Y_t^U \right]$ , using another SVR model as an ensemble tool.

**<Insert Fig.4 here>**

To summarize, the proposed BEMD-based SVR modeling framework can be abbreviated as a 'BEMD (decomposition)-SVR (single and ensemble forecasting)' hybrid model, based on the 'decomposition and ensemble' strategy.

In order to verify effectiveness of the proposed BEMD-SVR model, the classical EMD-SVR model employed as a univariate technique[25] and three well-established interval-valued time series methods, including interval Holt's exponential smoothing [17], vector error correction (VEC) model [19]and iMLP [18], are implemented for interval-valued electricity demand forecasting herein. The computational algorithm of these benchmark methods have been presented in many papers, so will not be repeated here to keep this paper concise. For detailed introduction to these methods, please refer to [17-19, 25].

## 4. Research design

This section provides details about the research design on data process and preliminary analysis, accuracy measure, methodologies implementation, and experimental procedure. The further experimental results and discussions are reported in the next section.

### 4.1 Data process and preliminary analysis

In the current study, interval-valued data do not come from noise assumptions, but rather from the aggregation of the high-frequency sample data in electricity market. First of all, we have to operate with the original time series in order to obtain the interval-valued time series. The available data represent the electricity power demand in Pennsylvania–New Jersey–Maryland (PJM) Interconnection in MWh per day and per hour form June 2, 2000 to December 31, 2011[1].

Following the procedure presented in [15], we calculate the maximum and minimum value of the demand per hour and month from 2000 to 2011, as shown in Fig. 2. The data are expressed in log scale. Note that the interval representation applied here is only the one formed by the lower and upper bounds. Accordingly, we have 24 interval-valued electricity demand series, one for each hour.

---

[1] Free data are available from the website of PJM Interconnection (http://www.pjm.com).

After constructing the interval-valued electricity demand series, the ITS per hour is split into an estimation sample and hold-out sample. The first 10 years of ITS from 2nd June 2000 to 31th December 2011 are used as estimation sample. The hold-out samples consist of the remaining data of last 24 months (two years) (to be forecast in a one-step-ahead fashion). Each examined model is implemented (or trained) on the estimation sample, and forecasts are produced for the whole of the hold-output sample. The forecasts are then compared to the hold-out sample to evaluate the performance of each model. Note that only one-step-ahead forecasting is considered in this study.

Here, we conduct the preliminary analysis using the example of monthly interval-valued electricity demand for H5 of year 2000 to 2009 as shown in Fig. 2. Looking at Fig. 2, it is obvious that the lower ($Y_t^L$) and upper ($Y_t^U$) bounds series move in tandem; neither series look to be stationary. Augmented Dickey-Fuller (ADF) results at 5% levels of significance (not reported here for brevity but available on request) confirm that these series are non-stationary in levels but stationary in first differences. These results call for a formal test of cointegration between $Y_t^L$ and $Y_t^U$. Thus, the Johansen test is used to investigate if there are any cointegrated relations between the variables. The Bayesian criterion is used to select the lag parameter $p$. According to both maximum eigenvalue and trace statistics, the null hypothesis of no cointegration is rejected at 5% levels of significance (see Table 2). Further, there is no evidence that there exists more than one cointegrating vector. We thus set the dimension of the cointegration space to 1, that is, the lower and upper bounds series of monthly electricity demand H5 of year 2000 to 2009 are considered to be CI(1,1). More importantly, Table 2 shows that when the coefficient of $Y_t^U$ is scaled to 1, the estimated cointegrating vector is (1, -0.99867), which is very close to (1, -1). In other words, the stationary error correction term is closely proxied by the monthly range variable $Y_t^R = Y_t^U - Y_t^L$ for H5. This results concur with the findings of He, Kwok [41], who detected the same conclusion when analyzing the daily lower and upper bounds series of the WTI (West Texas Intermediate) crude oil prices. The identical preliminary analysis mentioned above are conducted for the other interval-valued electricity demand series as well. These results suggest that the lower and upper bounds of monthly electricity demand each hour are integrated (not reported here for brevity but available on request).

**\<Insert Table 2 here\>**

## 4.2 Accuracy measures for ITS

In this paper, the performances of examined models are evaluated based on a well-known classical error measurement, which has been recently adapted in [17] for ITS problems: the interval U of Theil statistics ($U^I$), defined as

$$U^I = \sqrt{\frac{\sum_{j=1}^{n}\left(Y_{j+1}^U - \hat{Y}_{j+1}^U\right)^2 + \sum_{j=1}^{n}\left(Y_{j+1}^L - \hat{Y}_{j+1}^L\right)^2}{\sum_{j=1}^{n}\left(Y_{j+1}^U - Y_j^U\right)^2 + \sum_{j=1}^{n}\left(Y_{j+1}^L - Y_j^L\right)^2}} \qquad (7)$$

where $n$ denotes the number of fitted intervals, $\left[Y_t^L, Y_t^U\right]$ is the $t$th true interval, $\left[\hat{Y}_t^L, \hat{Y}_t^U\right]$ is the $t$th fitted interval.

The $U^I$ statistics is suitable for comparing the performances of a reference model with a naïve model (which assumes a random walk model), where the most recent interval is considered as the prediction of a future interval of the series, $\left[\hat{Y}_{t+1}^L, \hat{Y}_{t+1}^U\right] = \left[Y_t^L, Y_t^U\right]$. This statistics is less than one if the predictor performs better than a random walk model. However, it is greater than one if the predictor performs worse than a random walk model. A value of $U^I$ equal to 0 indicates perfect forecasting [17].

## 4.3 Methodologies implementation

Taking into account the amount of interval-valued time series (that is 24), it is necessary to estimate 24 models undertaking BEMD-SVR, interval Holt's exponential smoothing, vector error correction model, and iMLP respectively, one for each hour and 48 models undertaking EMD-SVR, as the EMD-SVR model is applied to independently forecast the lower and upper bounds of a given interval-valued electricity demand.

The proposed BEMD-SVR model is implemented in Matlab computing environment. Bivariate EMD[2] is implemented using the Matlab program provided by Rilling, Flandrin [31]. LibSVM[3] (version 2.86) [42] is employed for SVR modeling here. We select the radial basis function (RBF) as the kernel function in the BEMD-based prediction models when modeling the IMFs data. The linear kernel function is selected to model the residue and the relationship among the IMFs and the residue due to its simplicity and better performances after extensive experimental trials on different kernel functions. To determine the hyper-parameters, namely $C$, $\varepsilon$, $\gamma$ (in the case of RBF as the kernel function), straightforward grid search method is employed with exponentially growing grids of ($C$, $\varepsilon$, $\gamma$) parameters. Note that the fivefold cross validation is used in training phase to evaluate the modeling performance.

---

[2] Source code are available at http://perso.ens-lyon.fr/patrick.flandrin
[3] Source code are available at http://www.csie.ntu.edu.tw/~cjlin/libsvm/

For comparison purposes, as mentioned in Section 3, the classical EMD-SVR model, interval Holt's exponential smoothing, vector error correction model and iMLP are also implemented for interval-valued electricity demand forecasting. The implementations of these benchmarks are detailed as follows.

The classical EMD-SVR model is also implemented in Matlab computing environment. EMD[4] is implemented using the program provided by Wu and Huang [38]. The identical model selection procedure for SVR modeling in the BEMD-SVR model mentioned above is employed for the classical EMD-SVR model.

The interval Holt's exponential smoothing method (Holt[1]) is adopted here for interval-valued time series as it was done in [17]. The smoothing parameter matrix with elements constrained to the range (0, 1) can be estimated by minimizing the interval sum of squared one-step-ahead forecast errors. The solution of this optimization problem can be obtained using the limited memory BFGS method for bound constrained optimization (L-BFGS-B) which has been implemented in the R software package 'optimx'[5].

---

[4]  Source code are available at http://rcada.ncu.edu.tw/
[5]  R package 'optimx' are available at http://ftp.ctex.org/mirrors/CRAN/

The results of preliminary analysis presented in Section 4.1 suggest a cointegrating relationship between $Y_t^L$ and $Y_t^U$ with a cointegrating vector well approximated by (1, -1), a VEC model with $Y_t^R$ ($Y_t^R = Y_t^U - Y_t^L$) substituting for the error correction term is the natural empirical construct to examine their long-run and short-run interactions. The VEC model for monthly interval-valued electricity demand per hour is implemented using Eviews.

The iMLP is adopted here for interval-valued time series as it was done in [15, 18], but the source code is not free. Based on the formulation of iMLP presented by Roque, Maté [18], iMLP is implemented in Matlab computing environment. For minimizing the cost function formulated in [18], the BFGS quasi-Newton method and backpropagation procedure are applied. An iMLP with 15 neurons in hidden layer is trained with estimation sample. To prevent over-fitting, we use the common practice of fivefold cross validation in iMLP modeling.

## 4.4 Experimental procedure

The monthly interval-valued electricity demand per hour is split into the estimation sample and the hold-out sample firstly. Then, the optimal five examined models for estimation sample are determined. Afterwards, obtained five models are used for interval-valued electricity demand forecasting for hold-out sample and the accuracy measures are computed. We repeat the previous modeling process 100 times yielding 100 accuracy measures. Upon the termination of this loop, performances of the examined models are judged in terms of the mean of each accuracy measure of 100 replications for hold-out samples. Analysis of variance (ANOVA) test procedures are used to determine if the means of performance measures are significantly different among the five models for each hour. If so, Tukey's honesty significant difference (HSD) tests [43] are then used to further identify the significantly different prediction models in multiple pair-wise comparisons.

## 5. Results

This section demonstrates the usefulness of the models through statistical evaluations in experiments using interval-valued electricity demand series. The experiments are detailed below.

The aim of this evaluation is to highlight the advantages of using the proposed BEMD-SVR model (decomposing the interval bounds simultaneously and then fitting the resulting components independently) for forecasting interval-valued electricity demand, relative to the classical EMD-SVR model (decomposing the interval bounds independently and then fitting the resulting components independently) and the three well-established ITS forecasting models (fitting the interval bounds simultaneously), i.e., Holt[1], VEC, and iMLP.

The prediction performances of five examined models (BEMD-SVR, EMD-SVR, Holt[1], VEC, and iMLP) in terms of $U^I$ are shown in Table 3. Note that the columns labeled as 'Trans1' and 'Trans2' show that the prediction accuracy measure of BEMD-SVR using two transformations (Eq. (1) and Eq. (2)) mentioned in Section 2.2. For each row of the Table 3 below, the entry with the smallest value is set in boldface and marked with an asterisk, and the entry with second smallest value is heighted in bold. Several observations can be drawn from Table 3.

**<Insert Table 3 here>**

Overall, the top three models across 24 hours turn out to be BEMD-SVR, then iMLP, and then VEC. It is clear that the proposed BEMD-SVR model is with position within top two. Indeed, the BEMD-SVR produces the most accuracy forecasts at 20 hours and obtains the second best forecasts at the remaining 4 hours. It is conceivable that the reason for the superiority of the proposed BEMD-SVR model is that decomposition strategy does effectively improve prediction performance of ITS.

As far as the comparison between the proposed BEMD-SVR and classical EMD-SVR, the BEMD-SVR is consistently the best performing model for ITS forecasting, this attests to the value which is added by decomposing simultaneously both the lower and upper bounds of interval-valued electricity demand series using BEMD technique.

Concerning the comparison between the interval artificial intelligence models (i.e., BEMD-SVR model and iMLP) and the traditional statistic models (i.e., Holt[I] and VEC), it is clear that the BEMD-SVR and iMLP outperform the Holt[I] and VEC, particularly in the period from 10 to 24 h. As it is known, the hourly electricity power demand is influenced by various factors, such as temperature, day of the week, etc, exhibiting strong nonlinearity. Fig. 5 shows an hourly average demand curve comparing the summer period vs. the winter period for 2011. Looking at Fig. 5, it is clear that in the period from 10 to 24 h, the demand shows the highest variability, as it is behavior differs considerably from summer to winter due to the temperature effect. In contrast, the demand curve from 1 to 9 h is relatively stable in both seasons due to the majority of companies and factories are closed and most of people are asleep. According to the variability of the demand in both seasons and the superiority of the BEMD-SVR and iMLP from 10 to 24 h, thus, we can conclude that the interval artificial intelligence methods are more efficient than the traditional statistic methods for forecasting the interval-valued time series with highly variability.

**<Insert Fig.5 here>**

Concerning the comparison between the classical EMD-SVR model and the three well-established interval-valued forecasting models, we can see that, whatever the hours, the classical EMD-SVR model is outperformed by the three ITS forecasting models. It is conceivable that the reason for the inferiority of univariate technique (i.e., classical EMD-SVR model) for ITS forecasting is that the possible mutual dependency between the lower and upper bounds of ITS is ignored in nature.

As far as the comparison among the three selected interval-valued forecasting models, the VEC and iMLP almost achieve better accurate forecasts than Holt[I] over all 24 hours. Concerning the comparison between the iMLP and VEC, iMLP emerges the winner.

As was mentioned in Section 2.2, we attempt to compare two transformations (Eq. (1) and Eq. (2)) from empirical perspective in the context of interval forecasting of electricity demand. As far as the comparison between Eq. (1) and Eq. (2), the difference of prediction accuracy measure between these two transformations using BEMD-SVR is negligible.

An important criterion in interval forecasting of electricity demand is that the forecasted upper bound should be always greater than the forecasted lower bound. In this study, all prediction results meet this criterion. For the purpose of illustration, prediction results of monthly interval-valued electricity demand in H5 for hold-out sample is raised as an example. Fig. 6 shows difference value of forecasted interval (forecasted upper bound minus forecasted lower bound) of monthly interval-valued electricity demand in H5 for hold-out sample. Looking at Fig. 6, it is clear that all difference value are non-negative, indicating the forecasted upper bound is always greater than the forecasted lower bound.

**<Insert Fig.6 here>**

Following the experimental procedure mentioned in Section 4.4, an ANOVA procedure is performed to determine if there exists a statistically significant difference among the five models in the hold-out sample for each hour. The results are not included in detail to save space (available on request). All ANOVA results are significant at the 0.05 level, suggesting that there are significant differences among the five models. To further identify the significant difference between any two models, the Tukey's HSD test at the 0.05 level is used to compare all pair-wise differences simultaneously. Table 4 shows the results of the multiple comparison tests. For each hour, the models are ranked from 1 (the best) to 6 (the worst). Note that the entries labeled as 'BEMD-SVR (Trans1)' and 'BEMD-SVR-(Trans2)' mean the BEMD-SVR model using two transformations (Eq. (1) and Eq. (2)) mentioned in Section 2.2. According to the obtained results in Table 4, one can deduce the following observations. First, the BEMD-SVR significantly outperforms the counterparts for the overwhelming majority of hours. Second, the iMLP ranks the first at 2, 7, 11, and 15 h, but the superiority of iMLP at these hours is not significant. Third, the difference in prediction performance between iMLP and VEC is not significant (with the exception of the 4, 10, and 14 h). Fourth, the EMD-SVR performs significantly worse than the counterparts for the majority of hours. Fifth, the difference of prediction performance between two transformations (Eq. (1) and Eq. (2)) is not significant at 0.05 level. Finally, concerning the traditional statistic models, the difference in prediction performance between VEC and Holt[I] is not significant at the 0.05 level (with the exception of the 12 and 21 h).

**\<Insert Table 4 here\>**

Generally speaking, from the above experiments, we can draw the following five main conclusions. (1) The proposed BEMD-based SVR modeling framework is significantly superior to all other methods listed in the study (except for the 2, 4, 8, and 20 h), indicating that the strategy of 'decomposition and ensemble' can effectively improve prediction performance in the case of interval-valued electricity demand. (2) The interval-valued forecasting methods perform strikingly better than the univariate technique. (3) Although the classical EMD-based time series modeling framework, this is, EMD-SVR model here, has been recently well-established for single-valued time series forecasting, the main reason for the inferiority of classical EMD-SVR model for ITS forecasting is that the possible mutual dependency between the lower and upper bounds of ITS is ignored. (4) Due to the highly variability appearance of interval-valued electricity demand, interval artificial intelligence methods are more suitable for prediction than traditional statistic methods. Thus, this leads to the fifth conclusion. (5) The proposed BEMD-based SVR modeling framework can be used as a promising solution for interval forecasting of electricity demand.

# 6. Conclusions

Interval forecasting of electricity demand plays an increasingly important role in the electricity industry, as it could prove to be a potential tool for both power generators and consumers to make their plans. Following the philosophy of the 'divide and conquer', this study proposes a novel BEMD-based SVR modeling framework for monthly interval-valued electricity demand per hour forecasting in PJM Interconnection. The experimental study shows that the proposed modeling framework can improve prediction performance significantly and outperform statistically some well-established counterparts in terms of forecast accuracy measure and equality of accuracy of competing forecasts test. This indicates that the proposed BEMD-based SVR modeling framework is a promising tool for interval forecasting of electricity demand.

Besides electricity demand, the proposed BEMD-based SVR modeling framework might be used for other tough interval-valued time series forecasting task in energy market such as electricity price, which appeals further evidence. Furthermore, this study restricts attention exclusively to one-step-ahead forecasting, multi-step-ahead forecasts are of greater value to decision-makers than one-step-ahead forecasts in energy market. We will look into these issues in the future research.

## Acknowledgment

This work was supported by the Natural Science Foundation of China under Project No. 70771042, the Fundamental Research Funds for the Central Universities (2012QN208-HUST), the MOE (Ministry of Education in China) Project of Humanities and Social Science (Project No. 13YJA630002), and a grant from the Modern Information Management Research Center at Huazhong University of Science and Technology (2013WZ005 2012WJD002).

# Figures

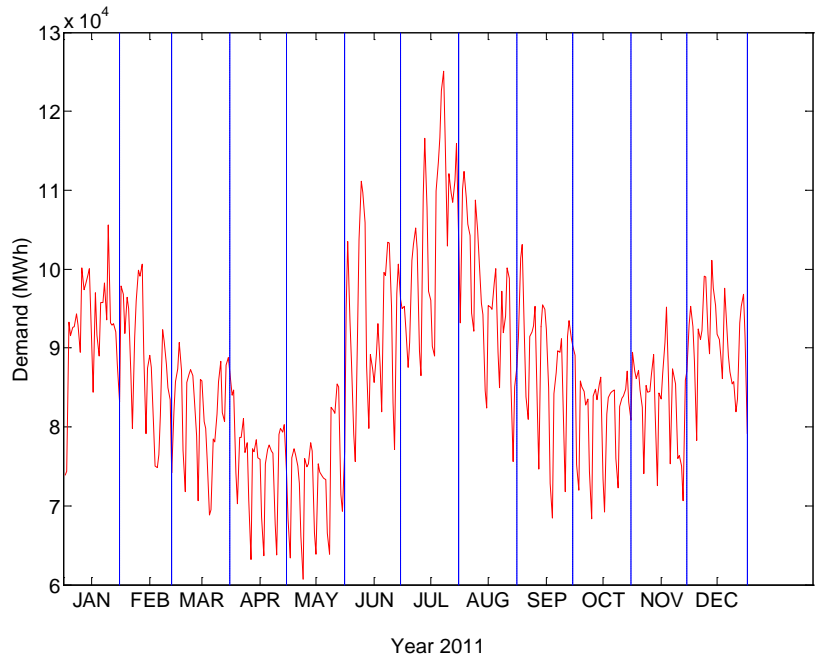

(a)

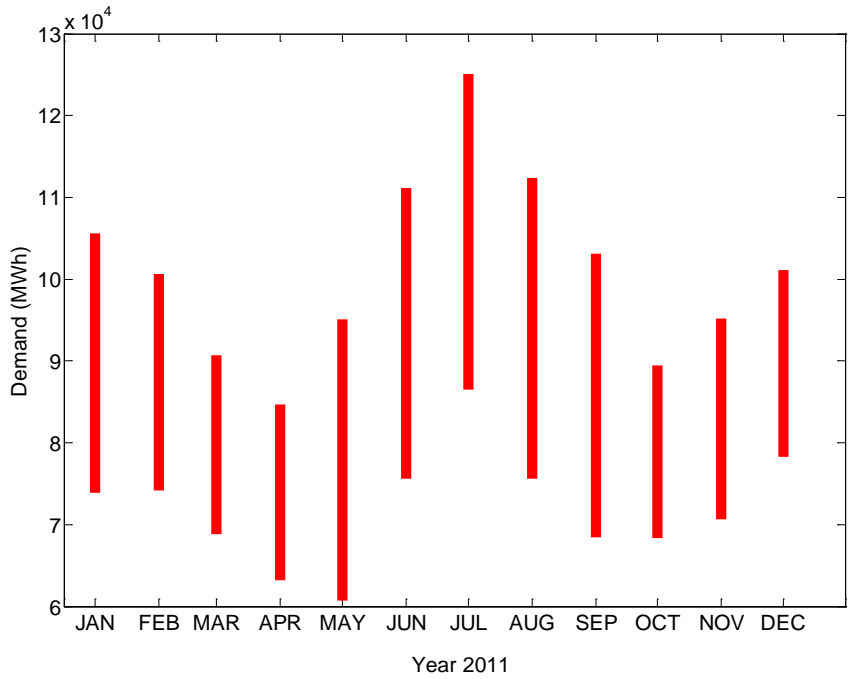

(b)

Fig. 1 (a) Daily electricity demand for H10, year 2011; (b) Corresponding monthly interval-valued electricity demand for H10, year 2011

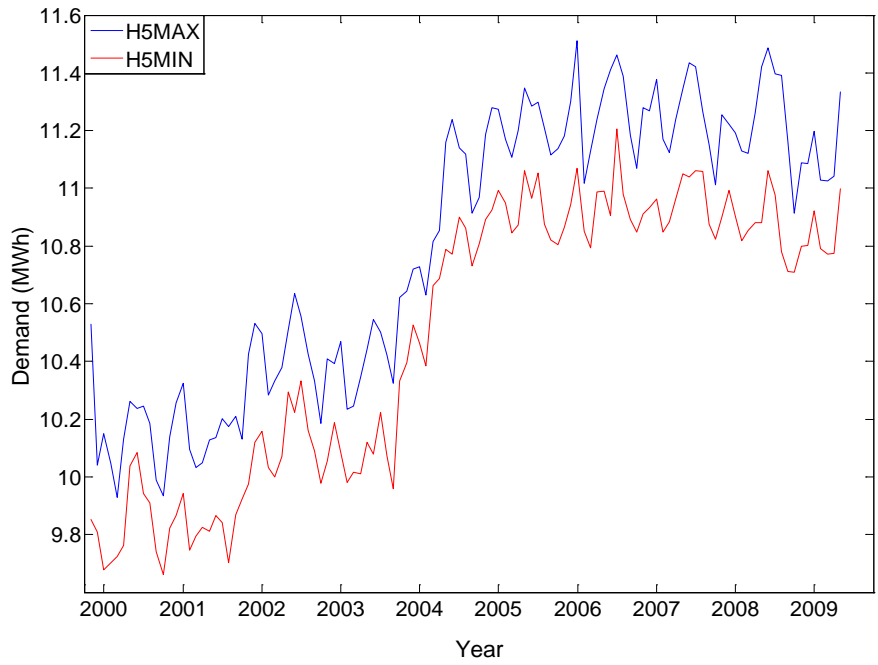

Fig. 2 Monthly interval-valued electricity demand for H5, year 2000-2009

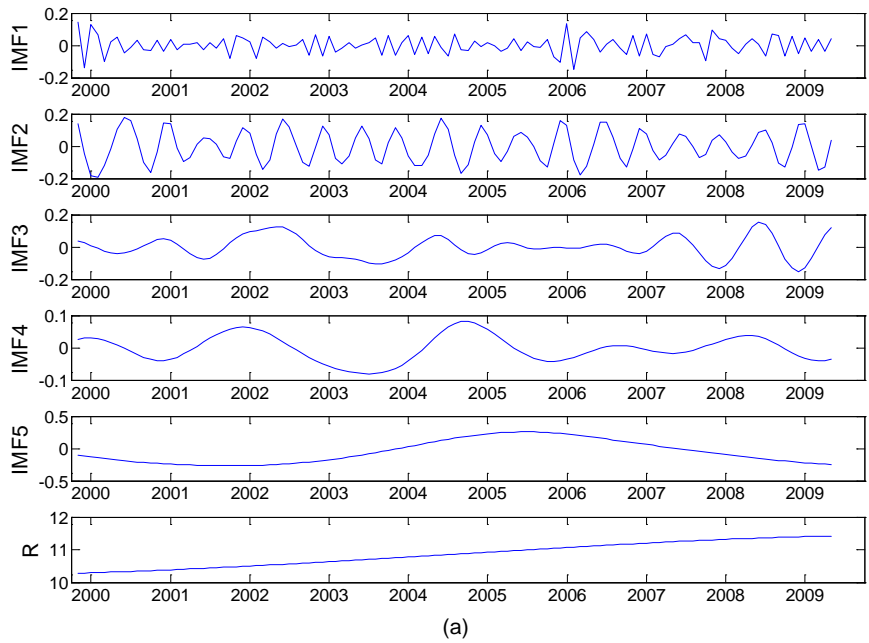

(a)

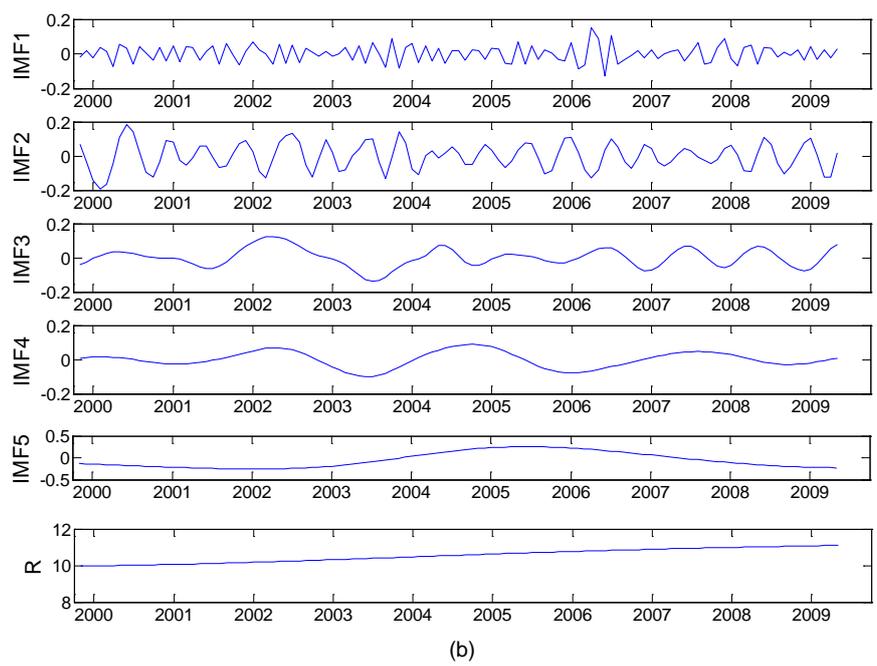

(b)

Fig. 3 The BEMD results of interval-valued electricity demand for H5, year 2000-2009: (a) Components for the upper bound series and (b) Components for the lower bound series.

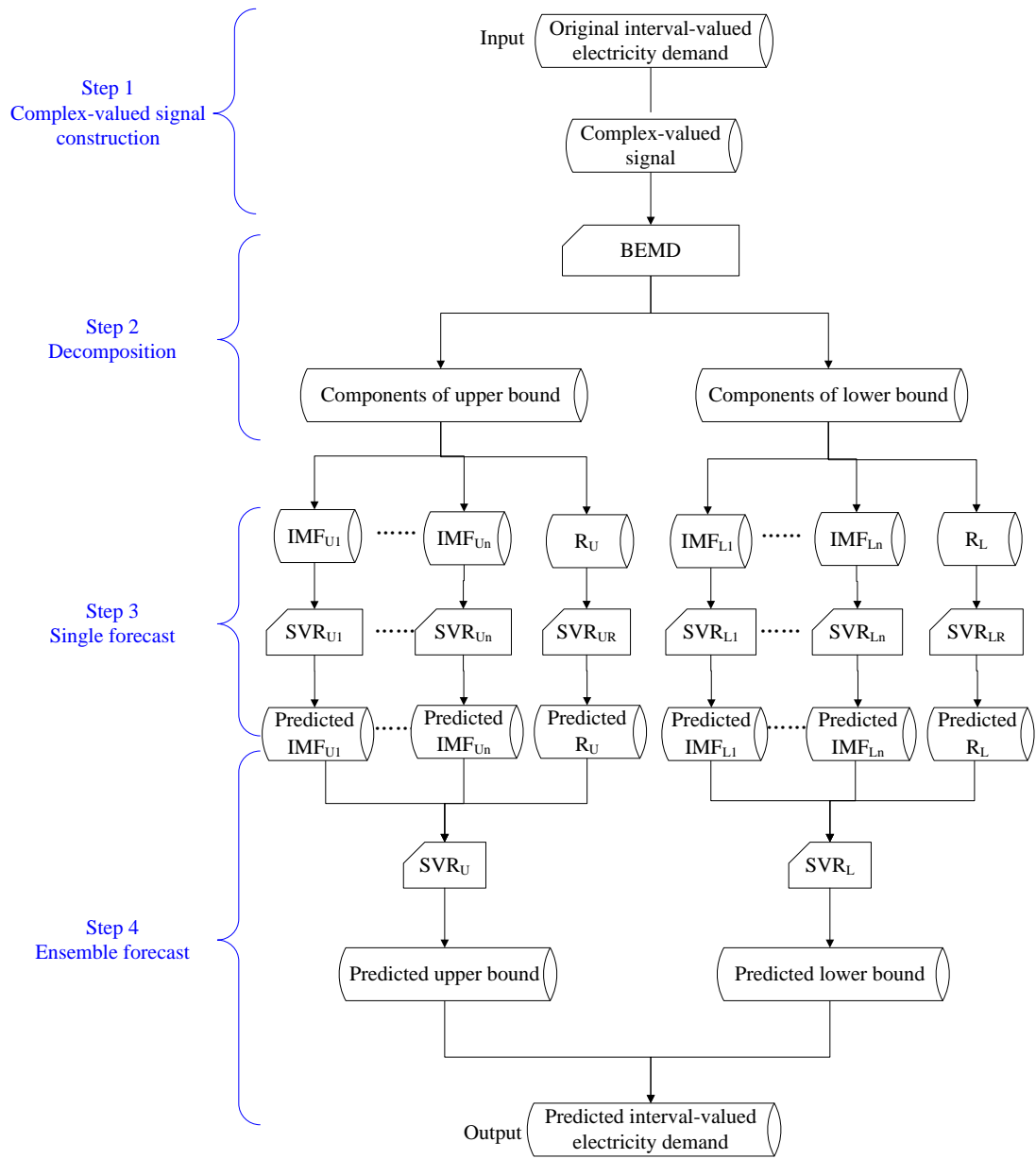

Fig. 4 General procedure of the proposed BEMD-based SVR modeling framework for interval-valued electricity demand forecasting

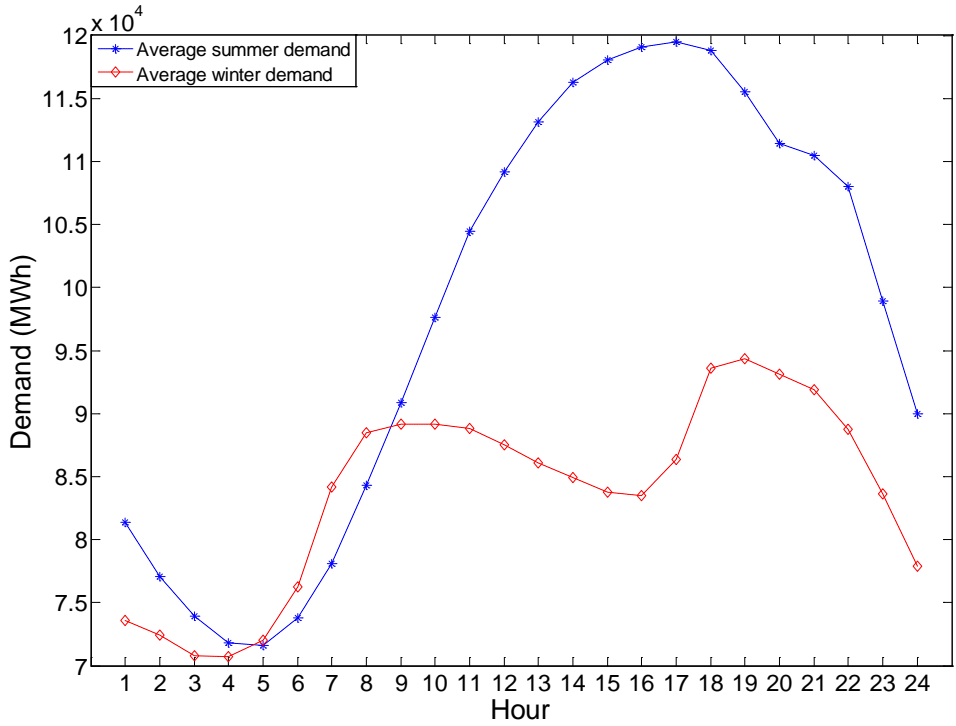

Fig. 5 Elapsed times of three models for each series of Mackey-Glass dataset

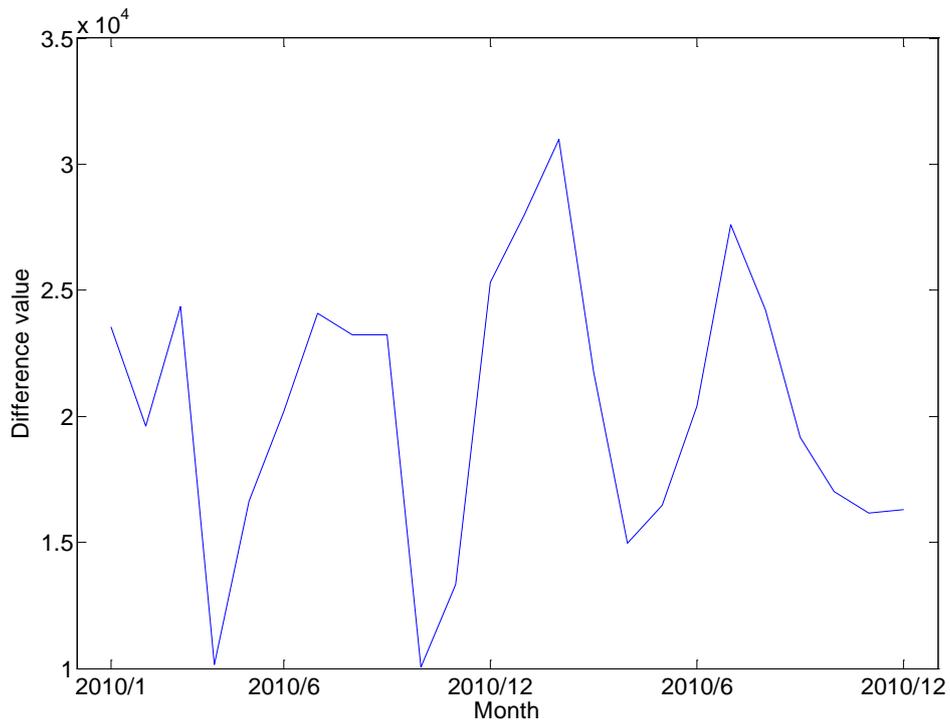

Fig. 6 The difference value of forecasted interval

# Tables

Table 1 Interval-valued variables

| Year 2011 | Electricity demand (MWh) | |
|---|---|---|
| | [Lower, Upper] | [Radius, Centre] |
| January | [64812, 89265] | [12226.5, 77038.5] |
| February | [59290, 88480] | [14595, 73885] |
| March | [56475, 73812] | [8668.5, 65143.5] |
| April | [53906, 65827] | [5960.5, 59866.5] |
| … | … | … |
| … | … | … |

Table 2 Cointegration test results

| | EIGENV | TRACE | U | L | LAG |
|---|---|---|---|---|---|
| r=1 | 0.011 | 1.498 | | | 5 |
| r=0 | 0.130** | 20.198** | | | 5 |
| Q(6) | | | 0.176 | 0.819 | |
| Q(12) | | | 4.519 | 8.374 | |
| C | | | 1 | -0.99867 | |

*Notes*: The results of test for cointegration between lower and upper bounds series of monthly interval-valued electricity demand for H5 of year 2000 to 2009. Eigenvalue and trace statistics are given under the columns 'EIGENV' and 'TRACE.' 'r=0' corresponds to the null hypothesis of no cointegration and 'r=1' corresponds to the hypothesis of one cointegration vector. The no-cointegration null is rejected and the hypothesis of one-cointegration vector is not rejected. 'U' and 'L' identify the Q-statistics associated with the monthly upper and lower bound series equations. All the Q-statistics are insignificant. The rows labeled 'C' give cointegrating vectors with the coefficient of the upper bound series normalized to one. 'LAG' the lag parameters used to conduct the test.

Table 3 Prediction accuracy measures ($U^I$) for hold-out sample

| Hour | $U^I$ measure | | | | | |
|------|------|------|---------|---------|------|------|
| | BEMD-SVR | | EMD-SVR | $Holt^I$ | VEC | iMLP |
| | Trans1 | Trans2 | | | | |
| 1 | **0.548\*** | **0.596** | 1.087 | 0.877 | 0.754 | 0.715 |
| 2 | 0.481 | **0.468** | 0.982 | 0.769 | 0.715 | **0.427\*** |
| 3 | **0.447\*** | **0.455** | 0.847 | 0.741 | 0.663 | 0.696 |
| 4 | **0.267\*** | **0.279** | 0.789 | 0.662 | 0.612 | 0.495 |
| 5 | **0.281\*** | **0.294** | 0.864 | 0.617 | 0.684 | 0.558 |
| 6 | **0.415** | **0.411\*** | 0.821 | 0.748 | 0.612 | 0.634 |
| 7 | 0.587 | 0.617 | 0.857 | 0.648 | 0.597 | **0.421\*** |
| 8 | **0.398\*** | **0.408** | 0.954 | 0.729 | 0.712 | 0.528 |
| 9 | **0.427** | **0.418\*** | 0.912 | 0.748 | 0.641 | 0.602 |
| 10 | **0.385\*** | **0.397** | 1.054 | 0.894 | 0.845 | 0.517 |
| 11 | 0.563 | **0.550** | 0.948 | 0.748 | 0.701 | **0.518\*** |
| 12 | **0.648\*** | **0.667** | 1.268 | 1.024 | 0.845 | 0.714 |
| 13 | **0.561\*** | **0.574** | 1.158 | 0.845 | 0.802 | 0.725 |
| 14 | **0.368\*** | **0.385** | 0.854 | 0.801 | 0.728 | 0.574 |
| 15 | 0.624 | **0.617** | 1.384 | 1.127 | 0.912 | **0.527\*** |
| 16 | 0.481 | **0.474\*** | 0.987 | 0.754 | 0.712 | 0.627 |
| 17 | **0.379** | **0.354\*** | 0.879 | 0.848 | 0.598 | 0.604 |
| 18 | **0.319\*** | **0.328** | 0.917 | 0.761 | 0.674 | 0.518 |
| 19 | 0.511 | **0.507\*** | 1.287 | 0.957 | 0.847 | 0.717 |
| 20 | 0.485 | **0.471\*** | 1.184 | 0.857 | 0.784 | 0.604 |
| 21 | **0.529\*** | **0.537** | 1.274 | 0.859 | 0.684 | 0.584 |
| 22 | 0.416 | **0.409\*** | 0.958 | 0.748 | 0.708 | 0.597 |
| 23 | **0.487\*** | **0.499** | 1.157 | 0.824 | 0.764 | 0.694 |
| 24 | **0.517\*** | **0.526** | 1.208 | 0.847 | 0.825 | 0.771 |

*Note*: For each row of the table, the entry with the smallest value is set in boldface and marked with an asterisk, and the entry with second smallest value is heighted in bold

Table 4 Multiple comparison results with ranked strategies for hold-out sample

| Hour($h$) | Rank of Models | | | | | | | | | | |
|---|---|---|---|---|---|---|---|---|---|---|---|
| | 1 | | 2 | | 3 | | 4 | | 5 | | 6 |
| 1, 5, 8, 13, 18, 23, 24 | BEMD-SVR (Trans1) | < | BEMD-SVR (Trans2) | <* | iMLP | < | VEC | < | Holt$^{\mathrm{I}}$ | <* | EMD-SVR |
| 9, 16, 19, 20, 22 | BEMD-SVR (Trans2) | < | BEMD-SVR (Trans1) | <* | iMLP | < | VEC | < | Holt$^{\mathrm{I}}$ | <* | EMD-SVR |
| 4, 10, 14 | BEMD-SVR (Trans1) | < | BEMD-SVR (Trans2) | <* | iMLP | <* | VEC | < | Holt$^{\mathrm{I}}$ | < | EMD-SVR |
| 6, 17 | BEMD-SVR (Trans2) | < | BEMD-SVR (Trans1) | <* | VEC | < | iMLP | < | Holt$^{\mathrm{I}}$ | < | EMD-SVR |
| 3 | BEMD-SVR (Trans1) | < | BEMD-SVR (Trans2) | <* | VEC | < | iMLP | < | Holt$^{\mathrm{I}}$ | < | EMD-SVR |
| 2, 11 | iMLP | < | BEMD-SVR (Trans2) | < | BEMD-SVR (Trans1) | < | VEC | < | Holt$^{\mathrm{I}}$ | <* | EMD-SVR |
| 7 | iMLP | < | BEMD-SVR (Trans1) | < | BEMD-SVR (Trans2) | < | VEC | < | Holt$^{\mathrm{I}}$ | <* | EMD-SVR |
| 12, 21 | BEMD-SVR (Trans1) | < | BEMD-SVR (Trans2) | < | iMLP | < | VEC | <* | Holt$^{\mathrm{I}}$ | < | EMD-SVR |
| 15 | iMLP | < | BEMD-SVR (Trans2) | < | BEMD-SVR (Trans1) | < | VEC | < | Holt$^{\mathrm{I}}$ | < | EMD-SVR |

*Note*: * indicates the mean difference between the two adjacent strategies is significant at the 0.05 level